# OVERHEAD-MNIST: MACHINE LEARNING BASELINES FOR IMAGE CLASSIFICATION

Erik Larsen[1], David Noever[2], Korey MacVittie[2], and John Lilly[2]

[1]PeopleTec, Inc., Austin, TX, USA
erik.larsen@peopletec.com
[2]PeopleTec, Inc., 4901 Corporate Drive. NW, Huntsville, AL, USA
david.noever@peopletec.com

## ABSTRACT

*Twenty-three machine learning algorithms were trained then scored to establish baseline comparison metrics and to select an image classification algorithm worthy of embedding into mission-critical satellite imaging systems. The Overhead-MNIST dataset is a collection of satellite images similar in style to the ubiquitous MNIST hand-written digits found in the machine learning literature. The CatBoost classifier, Light Gradient Boosting Machine, and Extreme Gradient Boosting models produced the highest accuracies, Areas Under the Curve (AUC), and F1 scores in a PyCaret general comparison. Separate evaluations showed that a deep convolutional architecture was the most promising. We present results for the overall best performing algorithm as a baseline for edge deployability and future performance improvement: a convolutional neural network (CNN) scoring 0.965 categorical accuracy on unseen test data.*

## 1. INTRODUCTION

This paper presents baseline metrics for image classification of 23 machine learning (ML) models trained on the Overhead-MNIST (OMNIST) dataset [1-6]. Image classification is a thrust of ML research that trains a computer algorithm to correctly identify subjects in never-before-seen images. Progress in model optimization requires comparison to known benchmarks. One popular dataset used for this purpose is the MNIST collection of handwritten numbers [7]. Similar in construction, OMNIST consists of satellite images – see Figure 1 – and can be used as a drop-in replacement for MNIST [1]. Each picture contains object variants from sparse edges to single, or multiple, instances of the indicated category. The sample images are accurately labelled, which provides a supervised learning dataset for small overhead imagery.

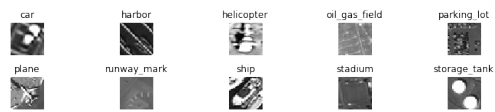

Figure 1. Samples from each class.

### 1.1. Background

Computer vision has now been the subject of ML research for decades, with pace and subject areas growing rapidly since the publication of Bengio, Bottou, Haffner, and LeCun in 1998 [9]. In more recent years the field of study has extended even into the realms of fashion [14] and molecular diffraction imaging and microscopy [15]. Alongside this content-space expansion has been the testing of numerous differing architectures attempting to improve on baseline performance. Chen et al. [16] randomly divided the original MNIST data into four different sub-groups and trained four popular models – CNN, ResNet, DenseNet, and CapsNet – in pursuit of advancing these benchmarks and gaining a better understanding of model performance on different sized groupings. More recently,

An et al. continued improving the baseline performance of CNN models with increasing complexity through robust hyperparameter tuning and augmentation of the training data [17]. We add to the increasing breadth of MNIST variants and baselines by examining Overhead-MNIST and presenting results for a variety of different ML model architectures.

## 1.2 O-MNIST Description

The Overhead-MNIST dataset is composed of 76,690 grayscale satellite images of shape (28, 28, 1), four comma-separated value (.csv) files containing either flattened picture arrays or label mapping/summary data, and ubyte file duplicates. There are 784 pixels per picture, and the raw arrays are not normalized. The ratio of test to train data is .125, while both share matching internal class distributions as seen in Figure 2. The full description of image sources and post-processing steps are described in [1].

The average train class size is 852, while the average test class size is 107. Helicopters bear the least representation with only 655 training and 82 test entries. This accounts for approximately 7.7% of the data, while classes for car, harbour, oil and gas field, parking lot, plane, ship, and storage tank compose around 10.4% on average. The remaining two classes, runway mark, and stadium, comprise 9.4% and 9.9% of the training data, respectively.

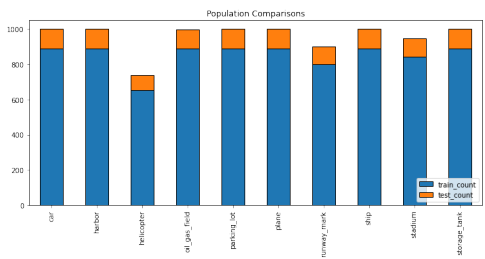

Figure 2. Training and test set size comparison.

The training and test sets share the same population distribution, with each showing significantly fewer helicopter examples. The runway_mark and stadium categories also have slightly less than average representation. While notable, these differences are unlikely to impact initial training or evaluation due to the relatively small discrepancy size. Attempts to project and cluster data into a 2-D surface with the UMAP algorithm [18] are shown in Figure 3.

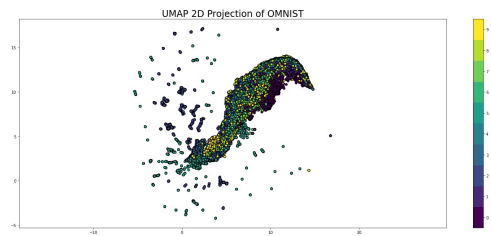

Figure 3. Poor separation of class clusters

## 2. METHODS

This research uses both Kaggle and Google Colab (Jupyter) notebooks with the Overhead-MNIST dataset found at Kaggle.com [2-6], and individual Python IDEs. ML libraries used include PyCaret, UMAP, sci-kit-learn, and TensorFlow's Keras API (2.4.0). Training and evaluation occur on a total of 23 classifiers [1,8-12]. The given test set remains unseen by the algorithm and is used for the final model scoring independent from PyCaret.

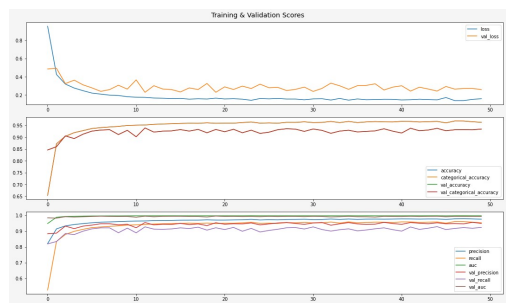

Figure 4. AlexNet architecture trained for 50 epochs with 0.3 dropouts and .00001 L1/L2 regularization in the fully connected layers.

## 2.1. Dataset Preparation

Raw data augmentation is limited to array normalization except for high contrast manipulations in CapsNet experimentation. Detailed image pre-processing is not evaluated. Normalization occurs during upload or model ingestion preparation via element-wise division by 255, the global pixel maximum. The presence of dissimilar class sizes necessitates stratification when further splitting the training set into train and validation sets; we use 20% of the training data for validation. The final results are based on predictions made from the unseen test data.

## 2.2. Validation and Scoring

We use categorical accuracy as the performance metric for comparison. Scoring is done internally with the PyCaret module using 10-fold stratified cross-validation which ranks by accuracy. Other metrics include precision, recall, F1 score, the area under the curve (AUC), Kappa, and MCC [12]. The eight best algorithmic models are shown below in Table 1, while score progression is seen in Figure 4.

CatBoost achieved the top scores from PyCaret but failed to break the 0.90 accuracy threshold. It also has the highest training time by a factor of almost three. Light Gradient Boosting achieves similar results but is still temporally expensive. The Extra Trees model gets only 3% less accuracy but is completed relatively quickly (100x speed advantage compared to CatBoost). Figure 5 shows the per-class accuracy as a confusion matrix, highlighting the relative misclassifications which are low for each satellite group.

Table 1. PyCaret Model Comparison Results

| Model | Accuracy | AUC | F1 | TT (Sec) |
|---|---|---|---|---|
| CatBoost Classifier | 0.8285 | 0.9808 | 0.8282 | 3391.5300 |
| Light Gradient Boosting Machine | 0.8001 | 0.9747 | 0.8001 | 101.1367 |
| Extreme Gradient Boosting | 0.7921 | 0.9736 | 0.7929 | 30.8533 |
| Extra Trees Classifier | 0.7314 | 0.9500 | 0.7297 | 30.9767 |
| Quadratic Discriminant Analysis | 0.7232 | 0.9363 | 0.7268 | 303.7533 |
| Gradient Boosting Classifier | 0.6621 | 0.9382 | 0.6634 | 3783.5900 |
| Random Forest Classifier | 0.5901 | 0.9103 | 0.5857 | 5.1400 |
| Logistic Regression | 0.4496 | 0.8369 | 0.4450 | 3.2133 |

*Note*: The full table of results can be found in the Kaggle notebook listed in References.

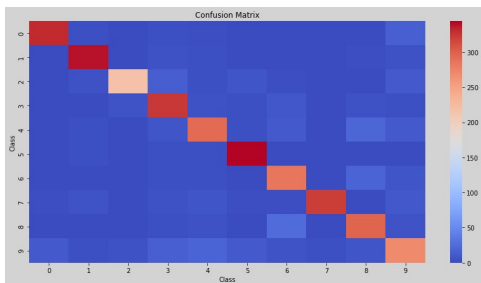

Figure 5. Heatmap for the CatBoost classifier test run confusion matrix.

To extend beyond the statistical machine learning approaches in Table 1, many convolutional neural networks (CNN) architectures were explored, including LeNet-5 [9], AlexNet [8], Graph Convolutional Network (GCN) [10], and CapsNet [11]. Likely due to the rigid, unaltered nature of the data (no crops or transformations), the GCN model performed poorly, reaching an accuracy of 46% on test data, and this only on the larger sets, as shown in Table 2.

GCNs rely on the division of the target images into subgraphs, which means that data that occupies the entirety of the frame proves difficult for the process to produce differentiable adjacency graphs, leading to poor performance [10]. While some classes were identifiable, the majority were not. Due to the averaging of pixel location to produce these graphs, attempting to use high-contrast data had no noticeable effect on the model's performance. Initial attempts to use the smaller dataset produced adjacency graphs that were almost completely identical.

A CapsNet model performs as well as the top models from the PyCaret comparison in Table 1, though specifically requiring a larger dataset. When trained on a smaller group it only resulted in 11% accuracy on testing, despite reporting 69.98% accuracy during its last training epoch. This is an indication of extreme overfitting that is likely insurmountable by structural or hyper-parameter means. The initial training reached 97.7% accuracy on its last epoch, but the test data again resulted in an 11% accuracy and overfitting to training inputs.

Table 2. CapsNet and GCN Results

| Model | Accuracy | Precision | Recall | F1 | TT (Sec) |
| --- | --- | --- | --- | --- | --- |
| CapsNet | .77 | .80 | .77 | .77 | 18325 |
| CapsNet (contrast) | .77 | .79 | .77 | .77 | 9328 |
| GCN | .46 | – | – | – | 1603 |

On larger data sets, however, the CapsNet model did perform well, hitting 77% accuracy, but only after a lengthy training period, roughly 15 minutes per epoch. This would be detrimental to situations requiring more immediate transfer weight updates, possibly risking strategic leverage. The initial model was trained for 20 epochs, while the model for the high-contrast transforms trained for only 10. The high-contrast model had no significant difference in performance. In training there were small accuracy gains with the high contrast data (90.36% accuracy reported at epoch 10, versus 89.01% at epoch 10 on the untransformed data).

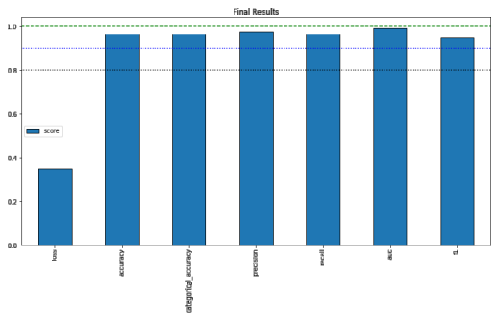

Figure 6. Best results for the tuned CNN model.

## 3. RESULTS

The best performing untuned classifier is a convolutional neural network (90% accuracy; see Figures 6-8), followed by Support Vector Machine with both accuracy and F1 scores of 73%. The Naïve Bayes and simple Multi-Layer Perceptron classifiers rank second and third, scoring 55% and 52%, respectively. All others fail to achieve 50% on average across the 10 classes, with a low of 9.5%. The K-Means algorithm tested is unsupervised, unlike the rest. Its overall performance is drastically lower, indicating a struggle to independently find separable pattern clustering.

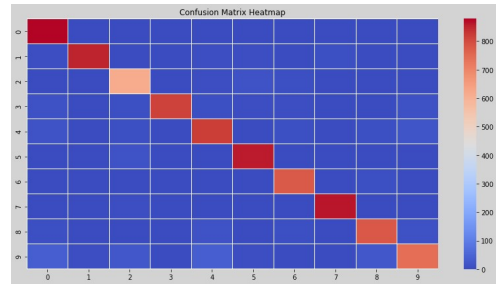

Figure 7. Heatmap of confusion matrix for the final CNN model.

While CapsNets are theoretically more robust than traditional convolutional networks (CNN) when it comes to certain machine vision tasks (e.g., facial recognition [13]) the nature of this dataset – in which the target can appear in random orientations [1] – combined with the small image size may have resulted in the capsule network's failing to recognize features consistent with specific target classes. It is possible that with additional training time, the network may have converged, but the final ten epochs of the unmodified data saw training accuracy scores all hovering around 90%, which indicates that the network was unable to learn more subtle nuances of the fitting function's parameters.

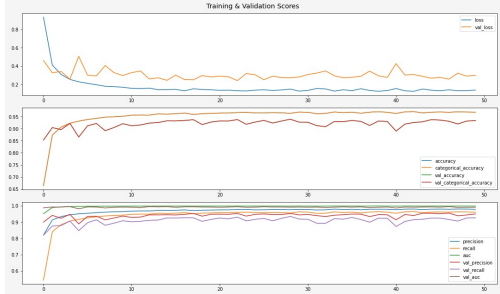

Figure 8. CNN model training progression for 50 epochs with dropout and regularization to mitigate overfitting.

## 4. CONCLUSIONS

A CNN model with hyperparameter tuning achieved test categorical accuracy of 0.9651 after only 500 training epochs, making it the choice for critical object classification applications. The model should be trained separately, and the updated weights deployed via transfer learning for on-board satellite classification. For active learning, new data could be constantly added to the database with the algorithm's predictions and then confirmed by human labelling. After verification, training should iterate with the new information shuffled into the bulk. Adding common image manipulation techniques such as random cropping, mirroring and rotations could also help improve classification accuracy.

The Overhead-MNIST satellite imagery dataset complements the ubiquitous MNIST handwriting example as a common starting point for algorithmic surveys. Models trained with OMNIST exhibit weaker performance than with their digital counterparts, presumably due to the large feature count and background clutter introduced through random, unrelated objects in the image coupled with a limited amount of un-altered images.

## ACKNOWLEDGMENTS

The authors would like to thank the PeopleTec Technical Fellows program for encouraging and assisting this research.

## Authors

Erik Larsen, M.S. is a senior data scientist with research experience in quantum physics and deep learning. He completed both M.S. and B.S. in Physics at the University of North Texas, and a B.S. in Professional Aeronautics from Embry-Riddle Aeronautical University while serving as an aviator in the U.S. Army.

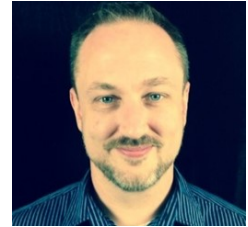

David Noever, Ph.D. has research experience with NASA and the Department of Defence in machine learning and data mining. Dr. Noever has published over 100 conference papers and refereed journal publications. He earned his Ph.D. in Theoretical Physics from Oxford University, as a Rhodes Scholar.

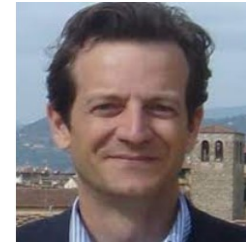

Korey MacVittie, M.S. is a data scientist specializing in machine learning. Prior research includes identifying undervalued players in sports drafting. He completed his M.S. at Southern Methodist University, and a B.S. in Computer Science, and a B.S. in Philosophy from the University of Wisconsin Green Bay.

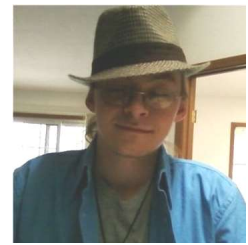

John R. Lilly III is an applied machine learning research scientist specializing in quantitative analysis, algorithm development, and implementation for embedded systems. Prior research includes cryptographic assets for multiple enterprise blockchain protocols. He has been trained in statistics by Cornell University. Formerly infantry, he currently serves as an intelligence advisor for the U.S. Army's Security Force Assistance Brigade.

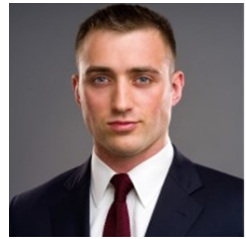